\documentclass{article}

\usepackage{tikz}
\usepackage{arxiv}
\usepackage{textcomp}
\usepackage{graphicx}
\usepackage[utf8]{inputenc} 
\usepackage[T1]{fontenc}    
\usepackage{hyperref}       
\usepackage{url}            
\usepackage{booktabs}       
\usepackage{amsfonts}       
\usepackage{nicefrac}       
\usepackage{microtype}      
\usepackage{lipsum}
\usepackage{multicol}

\title{A Fast Algorithm for Heart Disease Prediction using Bayesian Network Model}

\author{
 Mistura Muibideen and Rajesh Prasad\\
Department of Computer Science\\
African University of Science and Technology, Abuja, Nigeria\\
  \texttt{mmuibideen@aust.edu.ng, rprasad@aust.edu.ng} \\
 
}

\begin{document}
\maketitle\thispagestyle{empty}
\begin{multicols}{2}
\begin{abstract}
Cardiovascular disease is the number one cause of death all over the world. Data mining can help to retrieve valuable knowledge from available data from the health sector. It helps to train a model to predict patients’ health which will be faster as compared to clinical experimentation. Various implementation of machine learning algorithms such as Logistic Regression, K-Nearest Neighbor, Naïve Bayes (NB), Support Vector Machine, etc. have been applied on Cleveland heart datasets but there has been a limit to modeling using Bayesian Network (BN). This research applied BN modeling to discover the relationship between 14 relevant attributes of the Cleveland heart data collected from UCI repository. The aim is to check how the dependency between attributes affects the performance of the classifier. The BN produces a reliable and transparent graphical representation between the attributes with the ability to predict new scenarios. The model has an accuracy of 85\%. It was concluded that the model outperformed the NB classifier which has an accuracy of 80\%.
\end{abstract}

\keywords{: Naïve Bayes Classifier\and Bayesian network\and data mining\and Machine learning\and Artificial intelligence}

\section{Introduction}
The heart is a vital organ in the human body. It is essential for pumping blood in the circulatory system. The blood helps to convey oxygen which is needed for the functioning of the body cells. It has been established that the average heart pumps blood enough to fill a modern tanker annually and beats for about four million times annually [1].

Heart diseases are also known as cardiovascular diseases (CVDs). Heart diseases happen to be the most common cause of death globally. According to World Health Organization (WHO), both males and females are equally affected by heart disease. WHO estimated that 17.9 million people are dead from cardiovascular disease in 2016 which is about 31\% of all deaths worldwide. About 85\% of these global deaths are caused by stroke and heart attack [2].

Cardiovascular diseases result when there is a malfunction of the heart and blood vessels. One-fifth of deaths has been linked to cardiac and occurs when a trigger interacts with an arrhythmic substrate [3].

Some problems do exist along with cardiovascular diseases. There can be a case whereby the arteries harden and become inflexible and thicker. This is called arteriosclerosis. We can also have atherosclerosis which is the narrowing of the arteries that reduce the blood flow [4]. Heart attacks usually occur when there is a blood clot or blockage to the flow of blood from the heart. To buttress the importance of overcoming deaths of cardiovascular diseases, WHO launched a program on September 22, 2016 called the Global Hearts [5].

Some factors that tend to prone heart diseases are high blood pressure, smoking, high cholesterol, obesity, physical inactivity, unhealthy diet, and poorly controlled diabetes. Diagnosis of heart disease is usually done by taking of medical history, the use of a stethoscope, Ultrasound, and ECG.

Data mining helps to identify useful trends in a large set of data. The most important task in data mining is understanding and making discovery from data. It is a center of attraction to most scholars recently [6].  As a result of the increase in the amount of health data gathered through the electronic health record (EHR) systems, it is believed that strong analysis tools are necessary. With a huge amount of data, health care providers are now optimizing the efficiency of their organization using data mining. Data mining has helped the health care industry to specifically reduce costs by increasing efficiencies. It has helped to improve the quality of life of the patient, thereby saving the lives of patients. Data mining has been effective in predictive medicine, detection of fraud, customer relationship management, healthcare management and determining the effectiveness of some treatments [7]. Data mining can be used for different applications on health data. These applications have been roughly grouped into the following categories [8]:

Clinical decision making
Patients are normally examined by clinicians to detect their ailments. This process is experimental in nature and there is a possibility of the diagnosis being wrong. Data mining gives the clinician a second opinion for most diagnoses. This ensures that the disease is not under-estimated during diagnosis, thereby helping the clinicians in making better accurate predictions. 

Population health
Health analysts such as epidemiologists focus on the prevalence of diseases and are usually interested in finding trends, patterns, and causes of diseases across a population. Notable factors that they consider are early-life, lifestyle and social-demographic [8].

Health administration and policies
Handling insurance plans is a big challenge in health administration, especially when it comes to insurance fraud. Data mining has been successful in detecting insurance fraud in which the patients, doctors, or hospitals made claim for unnecessary drugs or procedures that did not actually occur. This may lead the insurance company to bankruptcy. The solution to this is a built predictive model that is real-time that helps in detecting the necessary drugs for each diagnosis. 

As a result of some risks identified with clinical treatments such as the delay in the result and the non-availability of the medical facilities to the people in the rural area, the prediction model is recommended. Although the prediction model is not an alternative to clinical treatment, it can serve as a first-hand tool to be aware of any type of disease and be prepared for it. There are many prediction systems that have been designed for different diseases diagnosis using different techniques. Examples of these diseases are breast cancer, diabetics, heart diseases, flu, cold, and uterine fibroid diseases, etc. Prediction of heart diseases has been going on for decades now. Most of the research has used different techniques such as Decision Tree, Naïve Bayes (NB), Support Vector Machine (SVM), and Neural Network, each showing different levels of accuracy. Some research used data set from the UCI repository while others use data from local source hospitals in their immediate environment.

Researchers in [9] used a multiclass performance classification SVM to diagnose the level of heart disease. The study used multiclass SVM algorithm namely: One-against-one (OAO), One-against-all (OAA), Binary Tree SVM (BTSVM), Decision Direct Acyclic Graph (DDAG) and Exhaustive Output Error Correction Code (EOECC). The dataset used was the UCI Cleveland dataset with BTSVM accuracy of 61.8\%. It was concluded that BT-SVM multiclass classification, OAO-SVM and SVM-DDAC provide better performance than the binary classification approach. Also, these algorithms provide a relatively stable performance for all levels. The occurrence of the imbalance dataset for sick-low, sick-medium, and sick-serious result in low performance of the system.

Researchers in [10] compares the different algorithm of decision tree classification for better performance in heart disease diagnosis. The algorithms tested are the J48 algorithm, Logistic model tree algorithm and Random Forest algorithm. Cleveland dataset from University of California, Irvine (UCI) data repository was used. It contains 303 instances and 76 attributes. The tool used for implementation was WEKA. J48 tree technique came out as the best classifier because it is more accurate and took the least time to build. This was followed by the logistic model tree and the Random Forest algorithm respectively. Application of reduced error pruning to J48 results in higher performance compared to where it was not applied. J48 has an accuracy of 56.76\% and build time of 0.04 seconds while the LMT algorithm has the lowest accuracy of 55.77\% and a build time of 0.39 seconds.

Authors in [11] used Naïve Bayes, K-Nearest Neighbor (KNN), Decision Tree and bagging technique. KNN was found to be the best technique with an accuracy of 79.2\%. WEKA tool was used for the implementation and the dataset used was Cleveland dataset.

Authors in [12] used a hybrid classification system based on Relief F and Rough Set (RFRS) method. The system is made up of two subsystems: the RFRS feature selection system and a classification system with an ensemble classifier. Data discretization, feature extraction and Relief F algorithm are the stages of the first system. An ensemble classifier is proposed for the second system. The dataset used is the stat log dataset obtained from the UCI repository. The accuracy was 92.59\% with MATLAB as the Implementation tool.

Authors in [13] used a number of algorithms including Decision Tree, J48 algorithm, Logistic Model Tree algorithm, Naïve Bayes, K-Nearest Neighbor (KNN) and SVM to predict heart diseases. They presented a new model that enhanced the decision tree accuracy in heart disease prediction. The WEKA and the UCI dataset were used in the implementation. Out of the four, the Naïve Bayes classifier was the best in performance followed by Support Vector Machine, the Decision Tree and then the KNN.

This research builds a probabilistic graphical model: Bayesian Network (BN) to understand the causal relationships among attributes of the Cleveland heart disease dataset from UCI. The BN produces a reliable and transparent graphical representation between the attributes with the ability to predict new scenarios which makes it an artificial intelligent tool. The model has been implemented and coded in R and Python with accuracy of 85\%. It was compared with NB classifier with accuracy of 80\%.

Parts of this paper are detailed as follows. Section 2 gives an insight of machine learning and concepts related to it. Section 3 discusses the research methodology used as well as the network design. The Cleveland heart disease dataset, the data preprocessing steps and the tools used for the study is also described in this section. Section 4 provides a detailed discussion on the results and system implementation. Section 5 rounds off the research by giving the conclusion.
 
\section{Related Concepts} \label{related-concepts}
This section describes some basic concepts and terminologies such as machine learning, classification and Bayesian network.
\subsection{Machine learning}

The field of Machine Learning (ML) has been in existence since 1959. Arthur Samuel while working for IBM defined ML as a field of study that enables the computer to learn without being explicitly programmed. A formal definition of ML was proposed by Tom Mitchell using a well-posed learning problem, stating that A is said to learn from experience E with respect to some task T and some performance measure P, if its performance on T as measured by P, improves with experience E.

To relate this definition to this research, we want to develop a heart disease prediction system. The task T of this system is to predict the presence of heart disease. The performance measure P is the prediction accuracy of our model. The system learns if we have more clinical data on heart disease status in patients. Here the experience E refers to the set of already processed clinical datasets. Hence as more records of data are added to the system, we achieve a higher precision as regards its accuracy. 

ML is a recurrent area of Artificial Intelligence (AI) with active research and applications in the past decades. Artificial Intelligence (AI) has been identified as one of the major role players in this current industrial revolution and there has been a lot of evolution in different machine learning algorithms [14]. Self-driven cars, speech recognition, robotic controls, effective web search, face detection are only a few of the areas where ML is being used.

\subsection{Naïve Bayes}
Naïve Bayes (NB) is the most commonly used classifier. There is a very strong independence assumption in a NB classifier. All the random variables are independent of each other given the class. Naïve Bayes is a machine learning algorithm. It has been identified as being effective in solving complex real-world problems [15].

Bayes theorem is used in calculating the posterior probability P(c | x), from the prior probability of predictor P(x), prior probability of class P(c) and the likelihood probability of predictor given class P(x|c)) [16].

\subsection{Bayesian Belief Networks}

It is a type of probabilistic graphical model. It is a directional acyclic graph made up of nodes and edges. The nodes denote the random variables while the edges stand for the causality. Each node has a conditional probability distribution (CPD) that shows the relationship of a node with its parents. The joint probability distribution is the product of several conditional distributions. The equation depicts the dependency structure by a direct acyclic graph. Where Pa (Xi) denotes the parent nodes of Xi. This equation is known as the chain rule [17]. 
    `           
The conditional independence assumption made by Naïve Bayes is sometimes rigid, especially in situations where the attributes are correlated. Hence the need for a more flexible way of modeling- Bayesian Belief Network (BBN). In BBN, we specify which pair of  attribute is conditionally independent. Bayesian network has been proved effective for uncertainty management in AI. It uses probability theory and graph theory to represent the relationship between nodes. BN modeling originated from data mining and machine learning. It reflects the probabilistic influences from big data sets [17]. 

\subsubsection{Some Basic Definition in Bayesian Belief Network}
Blocked

A path in between vertices A and B in a BN is blocked if it passes through a vertex C in a way that either:
\begin{itemize}
\item Serial Connection ((A\textrightarrow C\textrightarrow B) or ((A\textleftarrow C\textleftarrow B)) or diverging (A\textleftarrow C\textrightarrow B) and C is conditioned on
\item Converging (A\textrightarrow C\textleftarrow B) and neither C nor its descendants have received influence
\end{itemize}

D-separation

A and B are d-separated by C if all paths from a vertex of A to a vertex of B are blocked. If A and B are d-separated by C, then A is independent of B given C ($A\perp B\mid C$).

Active Trail

A trail is active, if there is a flow of influence from a node A to B through a node C. A causal trail (A\textrightarrow C\textrightarrow B), evidential trail (A\textleftarrow C\textrightarrow B), or common causal trail (A\textleftarrow C\textrightarrow B) is active only if C is not observed. A common effect trail (A\textrightarrow C\textleftarrow B) is active only if either C or one of C’s descendants is observed.

Markov’s Blanket

The Markov states that any node N is conditionally independent of another node given its Markov blanket. A node’s Markov blanket includes all its parents, children and children’s parents. Therefore, if a node is not present in the class attribute’s Markov blanket, its value is not relevant to the classification [18].

\subsubsection{Learning Bayesian network}
The two major operations in learning Bayesian network:

\begin{itemize}
\item Structural Learning 
\item Parametric Learning
\end{itemize}

Structure learning

This is the determination of the topology of the BN. Three popular structure learning algorithms are discussed below:

Constraint-based algorithms:

This analyzes the relations using the Markov property of BN with conditional independence tests. This can result in causal models even when learned from observational data. In this algorithm, an undirected graph could be generated and can be transformed into a BN using an additional independence test [17].

Score-based algorithms:

Algorithms assigns a number to each Bayesian network and then maximize it using some search algorithm, choosing the model with the highest score.

Hybrid algorithms:

This combines both the constraint and the score-based, it uses a conditional independence test to minimize the search space and the network score to identify the network with optimal performance.

Parametric learning 

This involves the calculation of the conditional probabilities in a network topology. Parameter learning is of two main types:

Maximum Likelihood Estimation (MLE):

This is the natural estimate for the CPDS. It simply uses the relative frequencies with which the variable states have occurred. According to MLE, we should fill the CPDs such that P(data | model) is maximized.

Bayesian Estimation:

The Bayesian Parameter estimator starts with the already existing prior conditional probability tables that express our beliefs about the variables before the data was observed.

\section{METHODOLOGY}
\label{Methodology}

This section discusses the various materials and methods in this study. The research design, Cleveland heart disease dataset, data preprocessing and the tools used for the study are discussed in detail.
\subsection{Network design}
\begin{figure*}

    \centering
    \includegraphics{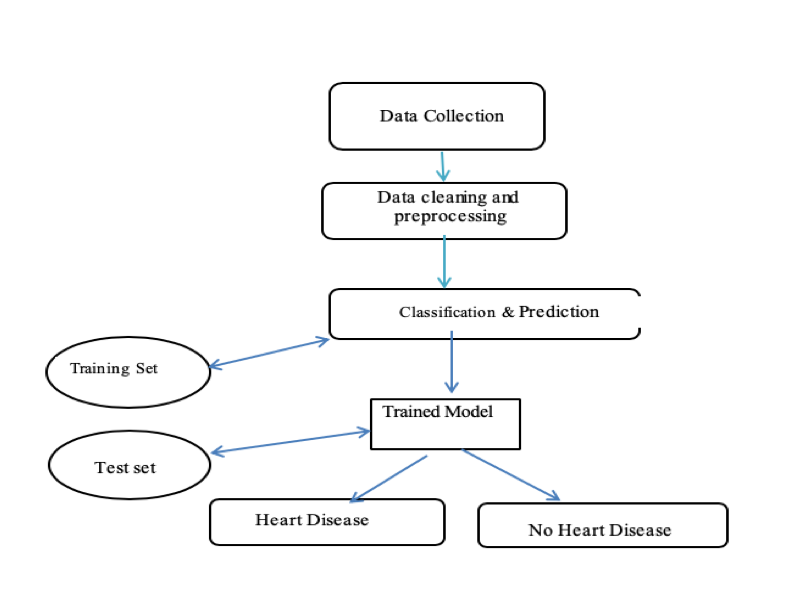}
    \caption{Flow Diagram of Network Design}
    \label{fig:fig1}
    
\end{figure*}

Network design describes the actual flow of the entire network building. A flowchart is shown in Figure \ref{fig:fig1} that explains the sequence involved in the network design.

\subsection{Cleveland Heart Disease Data set}
This research uses the heart disease dataset from the UCI machine learning repository. It consists of 303 samples with 14 attributes. Table \ref{tab:table1} shows the detailed description of the Cleveland dataset.

\begin{table*}
 \caption{Attributes of Cleveland heart disease dataset }
  \centering
   \begin{tabular}{lll}
    \toprule
     Attribute     & Description     & Domain of Values  \\
     \midrule
    Age 	&Age in years	&29-79    \\
    Sex  	&Sex	&0\textrightarrow Female, 1\textrightarrow Male \\
    Cp	    &Chest pain type	&1→Typical angina, 2→Atypical angina 3→Non-angina, 4→Asymptomatic\\
    Trestbps	&Resting blood sugar	&94 to 200 mm Hg\\
    Chol	&Serum cholesterol	&126 to 564 mg $dL^{-1}$\\
    Fbs	    &Fasting blood	sugar >120 mg $dL^{-1}$ & 0→ False, 1→True\\
    Restecg	   &Resting ECG result	&0→Normal, 1→ST-T wave abnormality2→LV hypertrophy\\
    Thalach	&Maximum heart rate achieved	&71 to 202\\
    Exang	&Exercise induced angina	&0→No, 1→Yes\\
    Oldpeak	&ST depression induced by exercise relative to rest	&0 to 6.2\\
    Slope	&Slope of peak exercise ST segment	&1→ Upsloping, 2→Flat, 3→Downsloping\\
    Ca	&Number of major vessels colored by fluoroscopy	&0-3\\
    Thal	&Defect type	&3→ Normal, 6→Fixed defect, 7→Reversible defect\\
    Target	&Heart diseases	&0-4\\
     \bottomrule
   \end{tabular}
  \label{tab:table1}
\end{table*}

\subsection{Preprocessing data}
Data preprocessing is also known as cleaning data. It is one of the most important steps to achieve the best from the dataset. This is a process whereby data inconsistencies such as missing values, out of range values, unformatted data, and noise are removed from the data. The process is usually time-consuming because it involves a lot of experimentation trying out various data analysis tools. Our preprocessing involves data retrieval, handling missing values, target class transformation and data discretization
\subsubsection{Data Retrieval}	
Data retrieval is usually the first step. Data  can be gotten from various sources. It can be as easy as someone handing over a file on a drive for you to analyze them directly. Or you need to download it or issue a database query to collect the data. Our dataset is downloaded from the UCI machine repository.
\subsubsection{Handling Missing Values}	
Missing data values is a common problem faced by analysts. This occurs due to different reasons such as incomplete extraction, corrupt data, failure to load the information, etc. This is a great challenge that must be fixed because good models are generated when you make the right decisions on how to fix it [19].
These are 5 ways of handling missing data:
i.	Deleting Rows
ii.	Replacing with mean/median/mode
iii.	Assigning a unique category
iv.	Predicting the missing values
v.	Using algorithms which supports missing values
We adopted Deleting rows since we have few missing values.
\subsubsection{Target Class Transformation}	
As stated in the data set description, the target class contains values (0, 1, 2, 3, 4). Where 0 means healthy (no heart disease) and (1, 2, 3, 4) means the presence of sickness of varying degrees. Interest is in the absence or presence of heart disease, so the need to limit the class to (0, 1). Level (1, 2, 3, 4) was converted to 1.
\subsubsection{Data Discretization}
In our dataset, 5 out of the 14 attributes are continuous. Variables in Bayesian network models are discrete in nature, and therefore we need to make this continuous data categorical. We rely on expert knowledge to discretize our data. The continuous attributes are age, trestsbp, chol, thalach, and oldpeak.
\subsection{Performance Metrics}
Performance metrics are used to evaluate how different algorithms perform based on various criteria such as accuracy, precision, recall etc. They are discussed below. 

Confusion Matrix\\
The confusion matrix shows the performance of the algorithm. It depicts how the classifier is confused while predicting. The rows indicate the actual instance of the class label while the columns indicate the predicted class instances. Table \ref{tab:table2} shows a confusion matrix for binary classification.

\begin{table*}
 \caption{ Confusion Matrix}
  \centering
  \begin{tabular}{lll}
    \toprule
     Actual Label    & Predicted Label\\
    \midrule
     & +(1)  & -(0)    \\
     \midrule
     +(1) &True Positive &False Negative\\
     \midrule
     -(0) &	False Positive	& True Negative\\
    \bottomrule
  \end{tabular}
  \label{tab:table2}
\end{table*}

True positive value signifies that the positive value is correctly predicted, false positive means the positive value is falsely classified, false negative means the negative value is falsely predicted while the true negative means the negative value is correctly classified. The Confusion matrix table is used to calculate different performance metrics as discussed below.

Accuracy\\
Accuracy is the ratio of the number of correctly classified instances to all the cases. It is the sum of TP and TN divided by the total number of instances.

Precision \\
Precision is the proportion of true positive instances that are classified as positive. It reflects the closeness of predicted values is to one another [20].

Recall \\
Recall is the proportion of positive instances that correctly classified as positive. Recall is known as sensitivity.

F1 Score \\
F1 score combines both precision and recall and finds a balance between both.

\section{RESULTS AND DISCUSSIONS}

This section discusses the results of the research and compares the Bayesian network model with the Naïve Bayes classifier.
\subsection{System Framework}
The model is derived using the bnlearn library in the R programming language. The bnlearn package was chosen because it generates model with easy to identify attributes. This model was then implemented in Python for learning, which is a powerful and relatively easy to use programming language. The Python version is 3.6.5, with libraries such as Pgmpy, Pandas, Numpy, Matplotlib, Scipy  and Scikit-learn. The machine used is MacBook with Processor 1.1GHz intel core m3 and 8GB memory. The dataset was split into training and test sets on a ratio of 80:20. 
\subsection{Structure Learning and Parameter Learning}
\subsubsection{Structure Learning}
\begin{figure*}

    \centering
    \includegraphics{{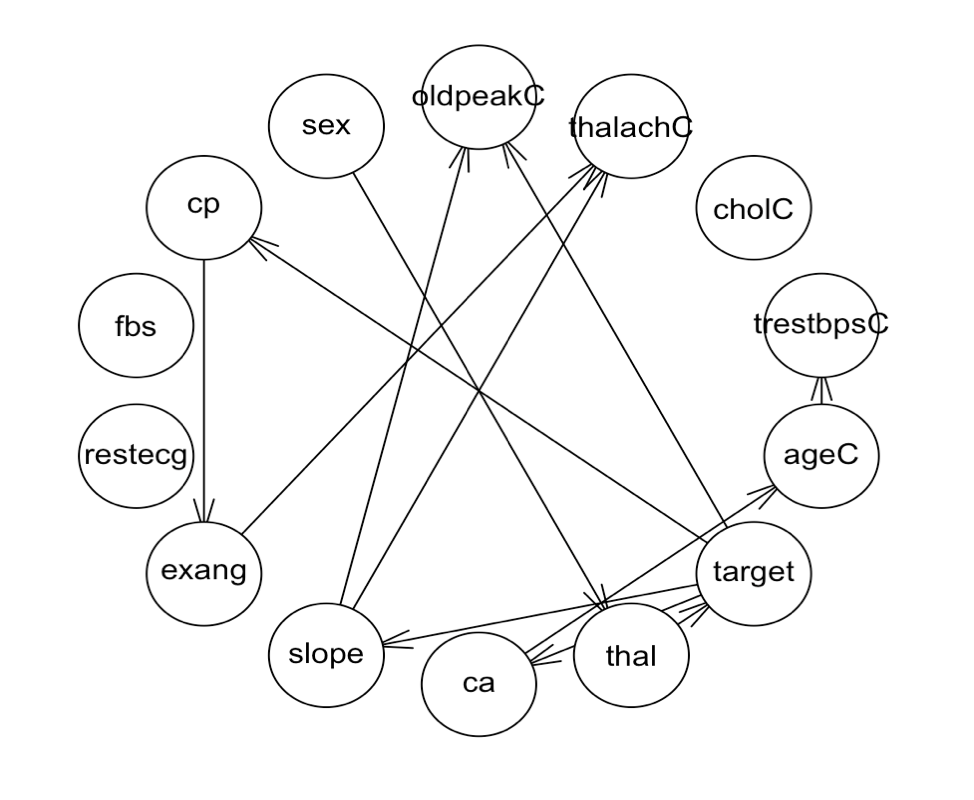}}
    \caption{Flow Diagram of Network Design}
    \label{fig:fig2}
    
\end{figure*}

After the preprocessing of our data to a form suitable to model, we use bnlearn in R to construct the belief network of the attributes. The generated model is shown in Figure \ref{fig:fig2}

\subsubsection{ParameterLearning}
The next thing is to learn the conditional probability table of each variable.

\textbf {Parameters of node sex, cp, fbs, restecg, exang, slope, ca, thal, target, ageC, trestbps, cholc}

Conditional probability table of attribute sex, cp, fbs, restecg, exang, slope, ca, thal, ageC, trestbps, cholc  as shown in Table \ref{tab:table3}. 

\begin{table*}
 \caption{Conditional Probability Table }
  \centering
  \begin{tabular}{lllllll}
    \toprule
    S/N	&Attributes	&  &0  &1 &2 &3 \\
      \midrule
    1	&sex  &		&0.3232323  &0.6767677\\
      \midrule
    2	&Cp|Target	&0	&0.10000000	&0.05109489\\
      \midrule
		& &1	&0.25000000	&0.06569343	\\
		  \midrule
		& &2	&0.40625000	&0.1313868\\
		  \midrule
		& &3	&0.24375000	&0.75182482\\
		  \midrule
	3	&fbs &		&0.8552189	&0.1447811\\		\midrule
    4	&Restecg &		&0.49494949	&0.01346801 &0.49158249\\
    \midrule
    5	&Exang |Cp	&0	&0.82608696	&0.91836735	&0.86746988	&0.45070423\\
    \midrule
		& &1	&0.17391304	&0.08163265	&0.13253012	&0.54929577\\
    \midrule
    6	&Slope|Target	&0	&0.64375000	&0.26277372\\	
		& &1	&0.30000000	&0.64963504\\
	\midrule
		& &2	&0.05625000	&0.08759124	\\
	\midrule
	7	&Ca|Target	&0	&0.8062500	&0.3284672\\		
		& &1	&0.1312500	&0.3211679\\
	\midrule
		& &2	&0.0437500	&0.2262774\\
	\midrule
		& &3	&0.0187500	&0.1240876\\
	\midrule
	
	8	&Thal|Sex	&0	&0.83333333	&0.41791045\\		
	 &	&1	&0.01041667	&0.08457711	\\
	 \midrule
	 
	&   &2	&0.15625000	&0.49751244	\\
	\midrule
	9	&Target|Thal	&0	&0.7743902	&0.3333333	&0.2347826\\
	\midrule
		& &1	&0.2256098	&0.6666667	&0.7652174\\
	\midrule
	
	10	&Agec |Ca	&0	&0.31034483	&0.07692308	&.02631579	&0.05000000\\
	\midrule
		& &1	&0.60344828	&0.75384615	&0.73684211	&0.65000000\\
	\midrule
		& &2	&0.08620690	&0.16923077	&0.23684211	&0.30000000\\
	\midrule
11	&trestbpsC|ageC	&0	&0.54098361	&0.25641026	&0.34146341	\\
    \midrule
		& &1	&0.39344262	&0.51794872	&0.21951220	\\
    \midrule
		& &2	&0.06557377	&0.22564103	&0.43902439\\
    \midrule
12	&cholC 	&	&0.1649832	&0.3265993	&0.5084175\\

    \bottomrule
  \end{tabular}
  \label{tab:table3}
\end{table*}

\textbf {Parameters of node thalachC}

Conditional probability table of attribute thalachC is as shown in Table \ref{tab:table4}. The table depicts P( thalachC|slope,exang).

\begin{table*}
 \caption{Conditional probability table of attribute thalachC }
  \centering
  \begin{tabular}{lllllll}
    \toprule
    slope = 0 \\
    \midrule
    exang	&0	&1\\
    \midrule
    thalachC\\
    \midrule
    0	&0.1504425	&0.3461538\\
    \midrule
    1	&0.8495575	&0.6538462\\
    \midrule
    slope = 1\\
    \midrule
    exang	&0	&1\\
    \midrule
    thalachC\\
    \midrule
    0	&0.4133333	&0.7580645\\
    \midrule
    1	&0.5866667	&0.2419355\\
    \midrule
    slope = 2\\
    \midrule
    exang	&0	&1\\
    \midrule
    thalachC\\
    \midrule
    0	&0.1666667	&0.7777778\\
    \midrule
    1	&0.8333333 	&0.2222222\\

    \bottomrule
  \end{tabular}
  \label{tab:table4}
\end{table*}

\textbf {Parameters of node oldpeakC}

Conditional probability table of attribute oldpeakC is as shown in Table \ref{tab:table5}. The table depicts P( oldpeakC|slope,target).

\begin{table*}
 \caption{Conditional probability table of attribute oldpeakC }
  \centering
  \begin{tabular}{lllllll}
    \toprule
    target = 0\\
    \midrule
    Slope	&0	&1	&2\\
    \midrule
    oldpeakC\\
    \midrule
    0	&0.990291262	&0.958333333	&0.555555556\\
    \midrule
    1	&0.009708738	&0.041666667	&0.444444444\\
    \midrule
    target = 1\\
    \midrule
    Slope	&0	&1	&2\\
    \midrule
    oldpeakC\\
    \midrule
    0	&0.944444444	&0.651685393	&0.166666667\\
    \midrule
    1	&0.055555556	&0.348314607	&0.833333333\\

    \bottomrule
  \end{tabular}
  \label{tab:table5}
\end{table*}

\subsection{ Performance Evaluation of the BN Model}

\begin{figure*}

    \centering
    \includegraphics{{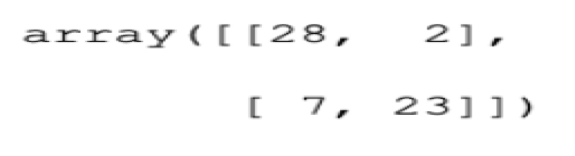}}
    \caption{Flow Diagram of Network Design}
    \label{fig:fig3}
    
\end{figure*}

\begin{figure*}

    \centering
    \includegraphics{{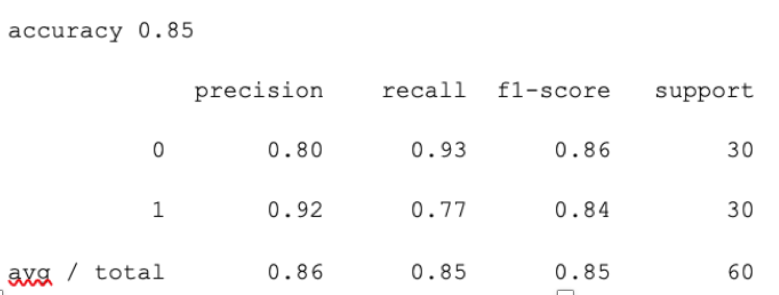}}
    \caption{Flow Diagram of Network Design}
    \label{fig:fig4}
    
\end{figure*}

This is to show how well our model performed. The model was able to predict 51 out of the 60 test samples correctly thereby achieving an accuracy of 85\%. The confusion matrix is as shown in Figure \ref{fig:fig3}. Performance metrics are as shown in Figure \ref{fig:fig4}

\subsection {Performance Evaluation of the Naïve Bayes Model}

The performance evaluation of the Naïve Bayes has an accuracy of 80\%. It was able to predict 49 out of the 60 test observations correctly. The confusion matrix is as shown in Fig 5. Performance metrics are as shown in Fig. 6.

\begin{figure*}

    \centering
    \includegraphics{{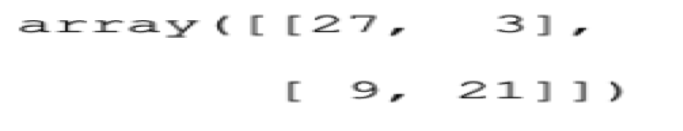}}
    \caption{Flow Diagram of Network Design}
    \label{fig:fig5}
    
\end{figure*}

\begin{figure*}
    \centering
    \includegraphics{{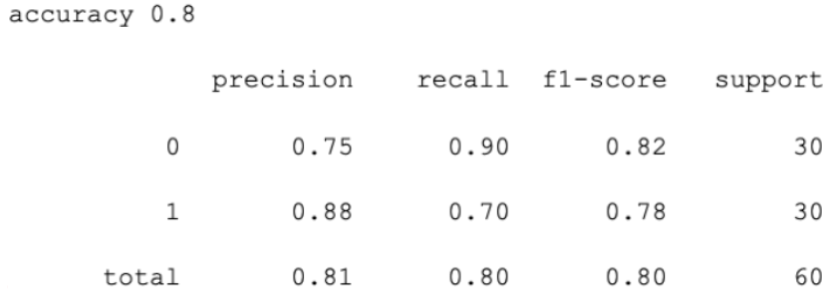}}
    \caption{Flow Diagram of Network Design}
    \label{fig:fig6}
    
\end{figure*}

\subsection{Discussion}

It is obvious from the performance metrics of both BN and Naïve Bayes that the BN outperformed the Naïve Bayes. The BN model was able to classify 85\% of the test dataset correctly compared to the 80\% achieved by the Naïve Bayes. It happens dues to the fact that the independence assumption made in Naïve Bayes affects the classifier accuracy. Here in BN, independent assumption between attributes has been relaxed. Other performance metrics such as precision, recall and f1-score also show that BN is better as shown in Fig.4 and Fig.6.

This result is logical as BN considers the dependency between attributes compared to the strong independence assumption made in Naïve Bayes that all random variables are independent of each other given the class output.
\section{CONCLUSION}

The research work developed a Bayesian network model for heart disease prediction in a human being. This model was built using the bnlearn package in R. The goal of this research is to compare the effectiveness of the Bayesian classifiers in predicting heart diseases. We used two different implementations of Bayesian classifier: the Bayesian Belief Network and the Naïve Bayes. The Bayesian Belief Network produced a graphical representation of the dependencies between attributes. The obtained model helps us to identify the causal dependencies and conditional independencies between attributes. 
Dataset was collected from the University of California, Irvine machine learning repository. It consists of 303 instances with 14 attributes; 13 numeric input attributes and one output. The Bayesian Belief Network Outperformed the Naïve Bayes in the prediction of heart diseases. This research will assist in making inferences about heart diseases, thereby serving as a diagnostic tool to support medical practitioners.  As a future work, our model can be compared with other classifier.

\end{multicols}
\end{document}